# Decomposition-Based Multi-Objective Evolutionary Algorithm Design under Two Algorithm Frameworks

**Lie Meng Pang[1], Member, IEEE, Hisao Ishibuchi[1], Fellow, IEEE and Ke Shang[1], Member, IEEE**

[1]Guangdong Provincial Key Laboratory of Brain-inspired Intelligent Computation, Department of Computer Science and Engineering, Southern University of Science and Technology, Shenzhen 518055, China.

Corresponding author: Hisao Ishibuchi (e-mail: hisao@ sustech.edu.cn).

This work was supported by National Natural Science Foundation of China (Grant No. 61876075), Guangdong Provincial Key Laboratory (Grant No. 2020B121201001), the Program for Guangdong Introducing Innovative and Enterpreneurial Teams (Grant No. 2017ZT07X386), Shenzhen Science and Technology Program (Grant No. KQTD2016112514355531), the Program for University Key Laboratory of Guangdong Province (Grant No. 2017KSYS008).

**ABSTRACT** The development of efficient and effective evolutionary multi-objective optimization (EMO) algorithms has been an active research topic in the evolutionary computation community. Over the years, many EMO algorithms have been proposed. The existing EMO algorithms are mainly developed based on the final population framework. In the final population framework, the final population of an EMO algorithm is presented to the decision maker. Thus, it is required that the final population produced by an EMO algorithm is a good solution set. Recently, the use of solution selection framework was suggested for the design of EMO algorithms. This framework has an unbounded external archive to store all the examined solutions. A pre-specified number of solutions are selected from the archive as the final solutions presented to the decision maker. When the solution selection framework is used, EMO algorithms can be designed in a more flexible manner since the final population is not necessarily to be a good solution set. In this paper, we examine the design of MOEA/D under these two frameworks. We use an offline genetic algorithm-based hyper-heuristic method to find the optimal configuration of MOEA/D in each framework. The DTLZ and WFG test suites and their minus versions are used in our experiments. The experimental results suggest the possibility that a more flexible, robust and high-performance MOEA/D algorithm can be obtained when the solution selection framework is used.

**INDEX TERMS** Evolutionary multi-objective optimization, MOEA/D, final population framework, solution selection framework, hyper-heuristics.

## I. INTRODUCTION

Multi-objective optimization problems are commonly found in many real-world applications [1]-[4]. The main goal of solving a multi-objective optimization problem is to optimize (either maximize or minimize) several objective functions simultaneously. However, the objective functions of a multi-objective optimization problem are conflicting in nature. Thus, it is not possible to obtain a single optimal solution that optimizes all the objective functions simultaneously. Usually a set of optimal solutions with different tradeoffs is obtained. The set of tradeoff optimal solutions is known as the Pareto optimal solution set. The Pareto optimal solutions form the Pareto front in the objective space.

Owing to the advantage of being able to search for a set of non-dominated solutions in a single run, evolutionary multi-objective optimization (EMO) algorithms have been a popular approach for solving multi-objective optimization problems. Over the years, various EMO algorithms have been proposed [5]. Since it is often impractical/difficult to include user preference a priori, most EMO algorithms in the literature are of the posteriori type [6]. That is, the main aim of these EMO algorithms is to find a set of well-distributed non-dominated solutions to approximate the Pareto front. Then, a decision maker chooses a solution from the obtained solution set.

Based on generation update mechanisms, EMO algorithms can be classified into two categories. One is non-elitist algorithms (e.g., MOGA [7] and NSGA [8]) and the other is elitist algorithms (e.g., SPEA [9] and NSGA-II [10]). In non-elitist algorithms, the current population is entirely replaced with the offspring population. No solution in the current population can survive to the next generation. As a result, it is very difficult to design an efficient EMO algorithm based on the non-elitist framework.

In elitist algorithms, good solutions (called elite individuals) in the current population survive to the next generation. The most frequently-used mechanism for generation update is the $(\mu + \lambda)$-selection strategy [10]. In this mechanism, $\lambda$ solutions are generated from the current population with $\mu$ solutions. Then, the best $\mu$ solutions are selected from the $(\mu + \lambda)$ solutions for the next generation. Some elitist EMO algorithms have an archive of a pre-specified size to store non-dominated solutions [11]-[13]. The archive is updated at every generation. In these EMO algorithms, the current population or the archive at the final generation is presented to the decision maker.

Even though the elitist framework is much more efficient than the non-elitist framework, it still has a difficulty. As discussed in [14], good solutions can be deleted during the generation update phase. Since the population size (and the archive size) is pre-specified, some solutions must be discarded during the generation update phase. New solutions cannot be compared with those discarded solutions in previous generations. As a result, solutions in the final population (and the final archive) are not always non-dominated among all the examined solutions [15].

In both the non-elitist and elitist frameworks, solutions in the final generation are presented to the decision maker. Thus, EMO algorithms should be designed to have a set of well-distributed non-dominated solutions in the final generation. In this paper, both the non-elitist and elitist frameworks are referred to as the *final population framework* (since most of recent algorithms are based on the $(\mu + \lambda)$ selection mechanism).

Recently, it was proposed to use an unbounded external archive in the design of EMO algorithms [14]. The idea is to present a solution set selected from all the examined solutions to the decision maker. In this paper, this algorithm framework is referred to as the *solution selection framework*. Whereas this framework was used for performance evaluation of existing EMO algorithms in some studies [16]-[17], it has not been used for the design of new EMO algorithms. Unlike the final population framework, the final population in the solution selection framework does not have to be a good solution set. The solution selection framework can use an EMO algorithm with a bounded internal archive together with an unbounded external archive (whereas we do not discuss such an EMO algorithm in this paper).

This paper aims to clearly demonstrate the flexibility of the solution selection framework in the design of EMO algorithms. We use Multi-objective Evolutionary Algorithm based on Decomposition (MOEA/D) [18] as a sample to empirically show the advantages of the solution selection framework. MOEA/D is a well-known and frequently-used EMO algorithm especially for many-objective optimization. A number of approaches have been proposed to further improve the performance of MOEA/D (e.g., with respect to the choice of a scalarizing function and the specification of weight vectors). In our former study [19], we demonstrated that the performance of MOEA/D is sensitive to the specification of the reference point and more robust performance is obtained by the solution selection framework.

Motivated by the promising experimental results in our former study about the reference point specification [19], we further investigate other components and parameters in MOEA/D in this paper. Our experiments are conducted using the two algorithm frameworks, as follows:

1.  **Final population framework**. The output of the MOEA/D algorithm is the solutions in the final population.
2.  **Solution selection framework**. The output of the MOEA/D algorithm is a solution set selected from all examined solutions in an unbounded external archive.

The examined MOEA/D components include a scalarizing function, a crossover operator, and a mutation operator. The examined parameters include the neighborhood size, the normalization parameter, the initial reference point and the final reference point in the dynamic reference point mechanism, the crossover rate, and the mutation rate. Since the parameter space is large and complex, we use an offline genetic algorithm-based hyper-heuristic method to search for the optimal configuration of MOEA/D under each algorithm framework for each test problem. Then, the obtained optimal algorithm configurations are compared and analyzed. As test problems, we use the three-objective DTLZ1-4 [20], WFG1-9 [21], and their minus versions [22]. We search for the best MOEA/D design for each of these 26 test problems under each algorithm framework.

This paper is organized as follows. First, we briefly explain multi-objective optimization and MOEA/D in Section II. Next, we explain our genetic algorithm-based hyper-heuristic method and the experiment settings in Section III. Then, we compare the experimental results under the two algorithm frameworks in Section IV. Finally, we conclude this paper in Section V.

## II. BACKGROUND

### A. MULTI-OBJECTIVE OPTIMIZATION PROBLEMS
Let us assume that we have the following *M*-objective minimization problem:

Minimize $\boldsymbol{f}(\boldsymbol{x}) = (f_1(\boldsymbol{x}), \ldots, f_M(\boldsymbol{x}))$ subject to $\boldsymbol{x} \in \mathbf{X}$, (1)

where $\boldsymbol{x} = (x_1, \ldots, x_D)$ is a *D*-dimensional solution vector, **X** is the feasible region, and $f_i(\boldsymbol{x})$ is the *i*-th objective to be minimized ($i = 1, 2, ..., M$).

***Definition 1 (Pareto dominance relation)***: A solution vector $\boldsymbol{x^A}$ is dominated by $\boldsymbol{x^B}$ if and only if $f_i(\boldsymbol{x^B}) \leq f_i(\boldsymbol{x^A})$ for all $i \in \{1, 2, ..., M\}$ and $f_j(\boldsymbol{x^B}) < f_j(\boldsymbol{x^A})$ for at least one index $j$.

***Definition 2 (Pareto optimal solution)***: A solution $\boldsymbol{x^*}$ is a Pareto optimal solution if and only if it is not dominated by any other solution in **X**.

The Pareto set (PS) consists of all Pareto optimal solutions. The projection of the Pareto set to the objective space is the Pareto front (PF).

### B. MOEA/D

The basic idea of MOEA/D [18] is to decompose a multi-objective optimization problem into $N$ single-objective subproblems using a set of predefined weight vectors $\boldsymbol{W} = \{\boldsymbol{w^1}, \boldsymbol{w^2}, ..., \boldsymbol{w^N}\}$ and a scalarizing function. Then, the $N$ subproblems are evolved in a cooperative manner by exploiting the information from other subproblems during the search process.

A set of uniformly distributed weight vectors $\boldsymbol{W}$ is used in most MOEA/D implementations in the literature. Each weight vector $\boldsymbol{w^j} = \left(w_1^j, w_2^j, ..., w_M^j\right)$ must fulfil the following relation:

$$w_1^j + w_2^j + \cdots + w_M^j = 1, \qquad (2)$$

where $w_i^j \geq 0$ ($i$=1, 2, ..., $M$) and $j \in \{1, 2, ..., N\}$. In our study, the Das and Dennis method [23] is used to systematically generate the weight vectors. Since each subproblem $j$ has a single individual $\boldsymbol{x^j}$ ($j \in \{1, 2, ..., N\}$), the population size equals to the number of subproblems (the number of weight vectors) in MOEA/D.

### C. COMPONENTS AND PARAMETERS IN MOEA/D

MOEA/D includes a number of components and parameters which need to be specified. They are explained in this subsection.

### 1. Scalarizing functions

It is known that a scalarizing function plays an essential role in MOEA/D. The scalarizing function is used to calculate the fitness value of each individual. In this paper, we consider five scalarizing functions: the weighted sum ($g^{\text{WS}}$), Tchebycheff ($g^{\text{TCH}}$), modified Tchebycheff ($g^{\text{MTCH}}$), penalty-based boundary intersection (PBI) ($g^{\text{PBI}}$), and inverted penalty-based boundary intersection (IPBI) ($g^{\text{IPBI}}$) functions. The five scalarizing functions are given as follows:

**Weighted Sum (WS) [18]:**

Minimize $g^{\text{WS}}(\boldsymbol{x}|\boldsymbol{w}) = w_1 f_1(\boldsymbol{x}) + \cdots + w_M f_M(\boldsymbol{x}),$ (3)

**Tchebycheff (TCH) [18]:**

Minimize $g^{\text{TCH}}(\boldsymbol{x}|\boldsymbol{w}, \boldsymbol{z^*}) = \max_{i=1,2,...,M}\{w_i \cdot |z_i^* - f_i(\boldsymbol{x})|\},$ (4)

where $\boldsymbol{z^*} = (z_1^*, z_2^*, ..., z_M^*)$ is the reference point (which is served as the origin of the weight vectors). In our study, we consider a dynamic reference point mechanism proposed in [24]. In principle, the ideal point is used as the reference point. However, the ideal point is unknown for real-world problems. Thus, the common practice is to specify each element $z_i^*$ of $\boldsymbol{z^*}$ by the minimum value of each objective $f_i(\boldsymbol{x})$ over all solutions examined so far, as shown in (5) and (6).

$$z_i^* = z_i^{min} - \epsilon_i, \ \ \epsilon_i \geq 0, \qquad (5)$$

$$z_i^{min} = \min\{f_i(\boldsymbol{x}), \boldsymbol{x} \in \mathbb{S}\}, i = 1, 2, ..., M, \qquad (6)$$

where $\epsilon_i$ is a non-negative number, and $\mathbb{S}$ consists of all examined solutions. The value of $\epsilon_i$ is set to zero in the original MOEA/D.

A dynamic reference point mechanism [24] has shown to be useful for improving the performance of MOEA/D. As discussed in [24], in the early stage of evolution, the estimated reference point is usually inaccurate since the population is not close to the Pareto front. The accuracy of the estimated reference point is gradually improved through the evolution. Based on these discussions, the following linearly decreasing formulation was proposed in [24]:

$$\epsilon_i = \left(\epsilon_i^{ini} - \epsilon_i^{end}\right)\left(\frac{T-t}{T-1}\right) + \epsilon_i^{end}, \qquad (7)$$

where $T$ is the maximum generation number, $t$ is the current generation index ($t = 1, 2, ..., T$), and $\epsilon_i^{ini}$ and $\epsilon_i^{end}$ are the initial and final settings of $\epsilon_i$, respectively.

**Modified Tchebycheff (MTCH) [25]:**

Minimize $g^{\text{MTCH}}(\boldsymbol{x}|\boldsymbol{w}, \boldsymbol{z^*}) = \max_{i=1,2,...,M}\{|z_i^* - f_i(\boldsymbol{x})|/w_i\},$ (8)

where $\boldsymbol{z^*}$ is the reference point. In this paper, we use the formulation in (5)-(7) as in the Tchebycheff function. If $w_i = 0$, $w_i$ is set to $10^{-6}$ to avoid division by zero.

**Penalty-based Boundary Intersection (PBI) [18]:**

Minimize $g^{\text{PBI}}(\boldsymbol{x}|\boldsymbol{w}, \boldsymbol{z^*}) = d_1 + \theta d_2,$ (9)

where the penalty parameter $\theta$ is a user-definable non-negative real number. The commonly-used value for $\theta$ is 5. The two distances $d_1$ and $d_2$ are defined as

$$d_1 = |(\boldsymbol{f}(\boldsymbol{x}) - \boldsymbol{z^*})^T \boldsymbol{w}|/\|\boldsymbol{w}\|, \qquad (10)$$

$$d_2 = \|\boldsymbol{f}(\boldsymbol{x}) - \boldsymbol{z^*} - d_1(\boldsymbol{w}/\|\boldsymbol{w}\|)\|, \qquad (11)$$

where $\boldsymbol{z^*}$ is the reference point specified by (5)-(7).

**Inverted PBI (IPBI) function [26]:**

Maximize $g^{\text{IPBI}}(\boldsymbol{x}|\boldsymbol{w}, \boldsymbol{z^N}) = d_1 - \theta d_2,$ (12)

where $\theta$ is the penalty parameter. The two distances $d_1$ and $d_2$ are defined as

$$d_1 = |(\boldsymbol{z^N} - \boldsymbol{f}(\boldsymbol{x}))^T \boldsymbol{w}|/\|\boldsymbol{w}\|, \qquad (13)$$

$$d_2 = \|\boldsymbol{z^N} - \boldsymbol{f}(\boldsymbol{x}) - d_1(\boldsymbol{w}/\|\boldsymbol{w}\|)\|, \qquad (14)$$

where $\boldsymbol{z^N} = (z_1^N, z_2^N, ..., z_M^N)$ is the estimated nadir point. Each element $z_i^N$ of $\boldsymbol{z^N}$ is specified by the maximum value of each objective $f_i(\boldsymbol{x})$ in the current population.

## 2. Normalization mechanism

In this paper, we consider a simple normalization mechanism [27] in MOEA/D to deal with problems with differently scaled objectives. The estimated ideal point $z^*$ (with $\epsilon_i = 0$ in (5)) and the estimated nadir point $z^N$ are used to normalize the objective value $z_i$ as follows:

$$z_i := \frac{z_i - z_i^*}{z_i^N - z_i^* + \varepsilon}, \qquad (15)$$

where $\varepsilon$ (which is referred to as the normalization parameter in this paper) is a positive real number used to prevent the denominator from becoming zero in the case of $z_i^N = z_i^*$.

## 3. Neighborhood structures

Neighborhood structure is an important feature of MOEA/D. Each subproblem has its own neighborhood which is defined by the Euclidean distance between weight vectors. For each subproblem, two parents are randomly selected from its neighborhood to generate a new solution. Then, the newly generated solution is compared with all solutions in its neighborhood (including the current solution of the current subproblem). All inferior neighbors are replaced with the newly generated solution. In the original MOEA/D, the same neighborhood is used for both mating and replacement.

In [28], the use of two different neighborhood structures for mating and replacement was investigated. For many-objective knapsack problems, it was shown that the search ability of MOEA/D can be improved by using a larger neighborhood structure for replacement than mating. In this paper, we also examine the use of two neighborhood structures.

## 4. Genetic operators

In the literature, many EMO algorithms use the SBX crossover [29] and the polynomial mutation [30] for multi-objective continuous optimization problems. In addition to these commonly-used genetic operators, we also examine some other operators: the whole arithmetic crossover (WAX) [31], the local arithmetic crossover (LAX) [32], the Gaussian mutation [30], and the random mutation [6].

## III. HYPER-HEURISTICS USING A GENETIC ALGORITHM

### A. ALGORITHM DESIGN SPACE FOR MOEA/D

As explained in the previous section, MOEA/D has a number of components and parameters to be specified. In this paper, we search for an optimal configuration of MOEA/D under each algorithm framework (i.e., final population and solution selection) for each test problem. More specifically, we try to find the best combination of the components and parameters in Table I. The domain of each component/parameter is shown in the column labeled "Domain" in Table I. For example, one of {WS, TCH, PBI, IPBI, MTCH} is selected as a scalarizing function. For the penalty parameter $\theta$ in PBI and IPBI, a real number is specified in the closed interval

[0, 10]. In this paper, the MOEA/D algorithm design means the choice of a possible value in the domain for each component/parameter in Table I.

In our experiments, the population size for MOEA/D is fixed to 91, and the distribution index for the SBX crossover and the polynomial mutation is fixed to their default value 20.

TABLE I.
PARAMETER SPACE FOR MOEA/D.

| Component/Parameter | Domain | Coding |
|---|---|---|
| $g$ (Scalarizing function) | {WS, TCH, PBI, IPBI, MTCH} | 3-bit |
| $\theta$ (only in PBI and IPBI) | [0,10] | 10-bit |
| $\epsilon_i^{ini}$ (Reference point) | {0, 0.001, 0.005, 0.01, 0.05, 0.1, 1, 2, 3, 4, 5, 6, 7, 8, 9, 10} | 4-bit |
| $\epsilon_i^{end}$ (Reference point) | {-5, -4, -3, -2, -1, -0.1, -0.05, -0.01, -0.005, -0.001, 0, 0.001, 0.005, 0.01, 0.05, 0.1} | 4-bit |
| $T_{Mate}$ (Mating neighbors) | {5%, 10%, 15%, 20%, 25%, 30%, 35%, 40%} | 3-bit |
| $T_{Rep}$ (Replacement neighbors) | {5%, 10%, 15%, 20%, 25%, 30%, 35%, 40%} | 3-bit |
| $\varepsilon$ (Normalization) | {0.000001, 0.00001, 0.0001, 0.001, 0.01, 0.1, 1, 5, 10, 15, 20, 25} | 10-bit |
| Crossover | {SBX, WAX, LAX} | 3-bit |
| $P_c$ (Crossover rate) | [0,1] | 5-bit |
| Mutation | {Polynomial, Gaussian, Random} | 3-bit |
| $P_m$ (Mutation rate) | [0,1] | 5-bit |

### B. GENETIC ALGORITHM-BASED HYPER-HEURISTIC

In this paper, we use a genetic algorithm-based hyper-heuristic method to find the optimal MOEA/D algorithm design for the final population framework and the solution selection framework.

We use the following parameter specifications in the hyper-heuristic method in our computational experiments:

Coding: 53-bit binary string,
Population size $\mu$: 100,
Initial population: Random binary strings,
Termination condition: 100 generations,
Crossover: Uniform crossover with probability 1,
Mutation: Bit-flip mutation with probability 1/53,
Selection: Tournament selection with tournament size 3,
Generation update: $(\mu + \mu)$-selection strategy,
Fitness evaluation: Average hypervolume.

In our hyper-heuristic method, each component/parameter is represented by a binary substring. For example, the scalarizing function $g$ is represented by a 3-bit substring $S_g = s_1 s_2 s_3$. It is decoded as

$$\text{Decode}(S_g) = 1 + \text{round}\left(\frac{4}{7}\sum_{i=1}^{3} s_i 2^{3-i}\right), \qquad (16)$$

where round($\cdot$) means round to the nearest integer. The WS, TCH, PBI, IPBI and MTCH functions are represented by the decoded values 1, 2, 3, 4, and 5, respectively. The penalty parameter $\theta$ for the PBI and IPBI functions is represented by a 10-bit substring $S_\theta = s_1 s_2 s_3 s_4 s_5 s_6 s_7 s_8 s_9 s_{10}$. The 10-bit substring $S_\theta$ is decoded as

$$\text{Decode}(S_\theta) = \frac{10}{1023} \sum_{i=1}^{10} s_i 2^{10-i}, \qquad (17)$$

where the 1024 real numbers in the interval [0, 10] are examined as the penalty parameter $\theta$. Then, an MOEA/D algorithm is represented by a binary string of length 53, where the coding for each component/parameter is listed in the third column of Table I.

The hypervolume (HV) indicator is used to evaluate each individual (i.e., the binary string of an MOEA/D configuration). The larger HV value indicates the better performance. In order to handle the stochastic nature of the MOEA/D algorithm, each MOEA/D configuration is applied to a test problem for five times. The average HV value over the five runs is used as the fitness value of each individual in the hyper-heuristic method.

In the final population framework, the HV value of the final population of each run of the MOEA/D configuration is calculated. In the solution selection framework, 91 solutions (which is the same as the population size of MOEA/D) are selected from non-dominated solutions among all the examined solutions in each run of the MOEA/D configuration. We use the distance-based solution subset selection method [33].

In the distance-based selection method, first, one of the extreme non-dominated solutions is randomly selected among the $M$ extreme non-dominated solutions as the first solution. The second solution is the non-dominated solution with the largest distance from the first solution. Then, the non-dominated solution with the largest distance from the first and second solutions is selected as the third solution. The selection process is repeated until 91 solutions are selected. We use this method since it is computationally efficient (e.g., it is fast). Of course, we can use various HV-based solution subset methods [34]-[36]. Whereas they can find better solution subsets with higher HV values, they need much more computation time than the distance-based method. In our hyper-heuristic method, 10,000 MOEA/D configurations (100 individuals × 100 generations) are applied to the given test problem five times. The solution subset selection is performed 50,000 times in a single run of the hyper-heuristic method. Thus, we use the fast distance-based solution subset selection method in the hyper-heuristic method in this paper. The termination condition for each MOEA/D configuration is 10,000 solution evaluations.

HV calculation needs a reference point. Since it is assumed that we have no knowledge about the true Pareto front of the test problem during the algorithm design by the hyper-heuristic method, we combined all solution sets obtained by the five runs of all MOEA/D configurations in the current population of the hyper-heuristic method. Then, we select only the non-dominated solutions among them to form an approximated Pareto front. The ideal point and the nadir point are calculated using the approximated Pareto front. The objective space is normalized so that the calculated ideal and nadir points become (0, 0, …, 0) and (1, 1, …, 1). Then, the reference point $r$ for HV calculation is specified as $r = (r, r, …, r)$ with $r = 1.1$ (i.e., a slightly worse point than the calculated nadir point in the normalized objective space).

## IV. EXPERIMENTAL RESULTS

### A. OBTAINED MOEA/D CONFIGURATIONS

Our computational experiments are performed for each of the three-objective DTLZ1-4 [20], WFG1-9 [21], and their minus versions [22] under each algorithm framework. The obtained MOEA/D configurations for the 26 test problems under the final population and solution selection frameworks are shown in Table II and Table III, respectively.

In Table II and Table III , we can see that the optimal MOEA/D configurations are totally different between the final population framework and the solution selection framework. As an example, when the final population framework is used, only the PBI and TCH functions are chosen as the scalarizing function for the DTLZ and WFG test suites (see Table II). However, when the solution selection framework is used, the WS function is selected for DTLZ2, DTLZ3, WFG2, WFG3, and WFG6 (see Table III).

Another example is the specification of the neighbourhood size. In the final population framework, the mating neighbourhood size ($T_{Mate}$) is always smaller than or equal to the replacement neighbourhood size ($T_{Rep}$). However, this is not always the case in the solution selection framework. As an example, for WFG9, the mating neighbourhood size (i.e., $T_{Mate} = 10\%$ of the population size) is larger than the replacement neighbourhood size (i.e., $T_{Rep} = 5\%$ of the population size).

### B. EVALUATION UNDER THE FINAL POPULATION FRAMEWORK

In this subsection, the performance of each MOEA/D configuration in Table II and Table III is evaluated under the final population framework. That is, the output of each algorithm is the solutions in the final population. For comparison purposes, we also perform the experiments for standard MOEA/D with the five scalarizing functions. For the standard MOEA/D, the following default parameters are used in the computational experiments:

Population size: 91,
Scalarizing functions: WS, TCH, MTCH, PBI, and IPBI,
Neighborhood size: 20,
Crossover: SBX with probability 1,
Distribution index for the SBX crossover: 20,
Mutation: Polynomial mutation with probability 1/$D$,
Distribution index for the polynomial mutation: 20.

TABLE II.
THE MOEA/D CONFIGURATIONS OBTAINED BY THE GENETIC ALGORITHM-BASED HYPER-HEURISTIC METHOD FOR EACH THREE-OBJECTIVE TEST PROBLEM UNDER THE FINAL POPULATION (FP) FRAMEWORK.

| Test problems ($M = 3$) | $g$ | $\theta$ (PBI/IPBI) | Reference Point $\epsilon_i^{ini}$ | Reference Point $\epsilon_i^{end}$ | $\varepsilon$ (normalization) | Neighborhood $T_{Mate}$ | Neighborhood $T_{Rep}$ | Crossover | $P_c$ | Mutation | $P_m$ |
|---|---|---|---|---|---|---|---|---|---|---|---|
| DTLZ1 | PBI | 1.2219 | 0.005 | 0.005 | 15 | 30% | 40% | SBX | 0.9677 | Polynomial | 0.1290 |
| DTLZ2 | PBI | 1.3099 | 0.001 | 0.005 | 1 | 15% | 35% | SBX | 1 | Polynomial | 0.0323 |
| DTLZ3 | PBI | 0.9971 | 6 | 0.001 | 20 | 40% | 40% | LAX | 0.4194 | Polynomial | 0.0968 |
| DTLZ4 | PBI | 1.5738 | 0.05 | 0.001 | 0.1 | 35% | 40% | SBX | 1 | Random | 0.0323 |
| WFG1 | TCH | - | 0.05 | 0.1 | 0.1 | 25% | 35% | SBX | 1 | Polynomial | 0.0645 |
| WFG2 | PBI | 3.4604 | 3 | 0.1 | 0.1 | 20% | 40% | SBX | 0.9677 | Polynomial | 0.1290 |
| WFG3 | TCH | - | 0.1 | 0.05 | 5 | 5% | 40% | - | 0 | Polynomial | 0.2903 |
| WFG4 | PBI | 1.3196 | 4 | -0.01 | 0.0001 | 35% | 40% | SBX | 1 | Polynomial | 0.0323 |
| WFG5 | PBI | 1.1437 | 3 | 0.005 | 0.001 | 35% | 40% | - | 0 | Polynomial | 0.0645 |
| WFG6 | PBI | 2.5611 | 1 | 0.005 | 10 | 40% | 40% | LAX | 0.7419 | Polynomial | 0.0323 |
| WFG7 | PBI | 2.0332 | 5 | 0.001 | 0.01 | 25% | 30% | LAX | 0.0968 | Polynomial | 0.0323 |
| WFG8 | PBI | 1.3294 | 1 | 0.05 | 0.000001 | 15% | 40% | SBX | 0.9677 | Polynomial | 0.0323 |
| WFG9 | TCH | - | 0.1 | 0.05 | 0.01 | 10% | 40% | LAX | 0.6129 | Polynomial | 0.2581 |
| Minus-DTLZ1 | IPBI | 1.2219 | 0.005 | -0.01 | 0.1 | 25% | 35% | SBX | 0.8710 | Random | 0.0323 |
| Minus-DTLZ2 | WS | - | - | - | 0.00001 | 35% | 40% | SBX | 1 | Polynomial | 0.0645 |
| Minus-DTLZ3 | WS | - | - | - | 25 | 35% | 40% | SBX | 1 | Random | 0.0323 |
| Minus-DTLZ4 | PBI | 0.1369 | 1 | 0 | 0.001 | 10% | 10% | LAX | 0.3548 | Gaussian | 0.5161 |
| Minus-WFG1 | IPBI | 2.5318 | 0.1 | 0.05 | 0.00001 | 10% | 35% | SBX | 0.8065 | Polynomial | 0.2581 |
| Minus-WFG2 | PBI | 1.7204 | 0 | 0.01 | 0.1 | 25% | 40% | SBX | 1 | Polynomial | 0.0323 |
| Minus-WFG3 | IPBI | 1.1046 | 0.05 | -0.05 | 0.0001 | 20% | 35% | SBX | 1 | Polynomial | 0.0323 |
| Minus-WFG4 | WS | - | - | - | 0.001 | 35% | 40% | SBX | 1 | Polynomial | 0.0645 |
| Minus-WFG5 | PBI | 0.1369 | 0 | -0.01 | 0.0001 | 40% | 40% | SBX | 1 | Polynomial | 0.0645 |
| Minus-WFG6 | WS | - | - | - | 0.001 | 40% | 40% | SBX | 1 | Polynomial | 0.0323 |
| Minus-WFG7 | WS | - | - | - | 0.001 | 35% | 40% | SBX | 1 | Polynomial | 0.0323 |
| Minus-WFG8 | PBI | 0.1564 | 9 | 0.005 | 0.001 | 15% | 40% | SBX | 1 | Polynomial | 0.2581 |
| Minus-WFG9 | WS | - | - | - | 0.0001 | 40% | 40% | SBX | 1 | Polynomial | 0.0645 |

TABLE III.
THE MOEA/D CONFIGURATIONS OBTAINED BY THE GENETIC ALGORITHM-BASED HYPER-HEURISTIC METHOD FOR EACH THREE-OBJECTIVE TEST PROBLEM UNDER THE SOLUTION SELECTION (SS) FRAMEWORK.

| Test problems ($M = 3$) | $g$ | $\theta$ (PBI/IPBI) | Reference Point $\epsilon_i^{ini}$ | Reference Point $\epsilon_i^{end}$ | $\varepsilon$ (normalization) | Neighborhood $T_{Mate}$ | Neighborhood $T_{Rep}$ | Crossover | $P_c$ | Mutation | $P_m$ |
|---|---|---|---|---|---|---|---|---|---|---|---|
| DTLZ1 | PBI | 0.5083 | 0.001 | 0.005 | 15 | 40% | 40% | SBX | 0.8387 | Polynomial | 0.1613 |
| DTLZ2 | WS | - | - | - | 20 | 5% | 35% | SBX | 0.4516 | Random | 0.0968 |
| DTLZ3 | WS | - | - | - | 0.001 | 40% | 40% | SBX | 0.9032 | Polynomial | 0.0968 |
| DTLZ4 | PBI | 2.2287 | 1 | -0.001 | 0.0001 | 40% | 40% | SBX | 1 | Random | 0.0323 |
| WFG1 | TCH | - | 0.01 | 0.05 | 15 | 30% | 40% | SBX | 1 | Polynomial | 0.0323 |
| WFG2 | WS | - | - | - | 15 | 5% | 35% | LAX | 0.9032 | Polynomial | 0.1613 |
| WFG3 | WS | - | - | - | 25 | 10% | 35% | LAX | 0.4194 | Polynomial | 0.1935 |
| WFG4 | TCH | - | 3 | -0.05 | 0.00001 | 40% | 40% | SBX | 1 | Polynomial | 0.0323 |
| WFG5 | PBI | 9.6090 | 8 | 0.005 | 0.0001 | 40% | 40% | SBX | 1 | Polynomial | 0.0645 |
| WFG6 | WS | - | - | - | 15 | 25% | 30% | LAX | 0.6774 | Polynomial | 0.0323 |
| WFG7 | PBI | 1.7791 | 6 | 0.1 | 1 | 10% | 40% | LAX | 0.5161 | Polynomial | 0.0323 |
| WFG8 | TCH | - | 0.1 | -0.005 | 25 | 10% | 35% | LAX | 0.9355 | Polynomial | 0.0968 |
| WFG9 | PBI | 0.8309 | 0 | 0.1 | 25 | 10% | 5% | LAX | 0.6129 | Polynomial | 0.0968 |
| Minus-DTLZ1 | IPBI | 1.9453 | 1 | -0.05 | 0.00001 | 30% | 40% | SBX | 1 | Random | 0.0323 |
| Minus-DTLZ2 | IPBI | 6.2757 | 7 | -0.001 | 15 | 40% | 35% | LAX | 0.2581 | Gaussian | 0.5484 |
| Minus-DTLZ3 | PBI | 0.2151 | 6 | -0.01 | 0.1 | 30% | 35% | SBX | 0.9355 | Random | 0.0323 |
| Minus-DTLZ4 | WS | - | - | - | 10 | 40% | 30% | LAX | 0.6774 | Gaussian | 0.8387 |
| Minus-WFG1 | TCH | - | 0.1 | 0.1 | 0.0001 | 30% | 40% | SBX | 1 | Polynomial | 0.0645 |
| Minus-WFG2 | MTCH | - | 1 | -0.1 | 0.01 | 30% | 30% | SBX | 1 | Polynomial | 0.0323 |
| Minus-WFG3 | PBI | 0.8602 | 4 | -1 | 0.1 | 35% | 35% | SBX | 0.9355 | Polynomial | 0.0323 |
| Minus-WFG4 | PBI | 0.3128 | 5 | 0 | 0.0001 | 30% | 30% | SBX | 0.8710 | Polynomial | 0.0645 |
| Minus-WFG5 | WS | - | - | - | 0.00001 | 40% | 40% | SBX | 1 | Polynomial | 0.0323 |
| Minus-WFG6 | PBI | 0.2151 | 10 | 0.005 | 0.0001 | 20% | 35% | SBX | 0.9677 | Polynomial | 0.0323 |
| Minus-WFG7 | TCH | - | 6 | -1 | 0.0001 | 30% | 40% | LAX | 0.2581 | Gaussian | 0.9032 |
| Minus-WFG8 | WS | - | - | - | 5 | 20% | 35% | SBX | 1 | Gaussian | 0.5806 |
| Minus-WFG9 | WS | - | - | - | 0.01 | 40% | 40% | SBX | 1 | Polynomial | 0.0323 |

As in [18], [22], and [26], the penalty parameter $\theta$ in the standard MOEA/D-PBI and MOEA/D-IPBI are set as 5 and 0.1, respectively. Our computational experiments are performed on the PlatEMO platform [37]. Each MOEA/D algorithm is independently run 31 times on each test problem.

In order to evaluate the performance of the MOEA/D, the HV indicator and the inverted generational distance (IGD) indicator are used. A larger HV value and a smaller IGD value indicate the algorithm has better performance. It should be noted that the HV indicator is also used to evaluate each MOEA/D configuration in the hyper-heuristic method (in Section III.B).

As we have explained in the previous section, a reference point is needed for calculating the HV value. In this section, the true Pareto front information of each test problem is used for evaluating the performance of MOEA/D. The true ideal and nadir points are used to normalize the objective space. The reference point $r = (1.1, 1.1, 1.1)$ is used for the hypervolume calculation. For the IGD calculation, a reference point set is needed. In our experiments, about 10,000 reference points are uniformly sampled over the entire Pareto front of each test problem in PlatEMO [37] as the reference point set for the IGD calculation except for WFG3 (see the footnote of Table IV).

Table IV shows the mean hypervolume value over 31 runs of each MOEA/D configuration for the three-objective test problems. The stopping condition for each MOEA/D configuration is 10,000 solution evaluations. We use the terms "MOEA/D-WS", "MOEA/D-TCH", "MOEA/D-MTCH", "MOEA/D-PBI", and "MOEA/D-IPBI" to denote the standard MOEA/D with the five scalarizing functions, respectively. The terms "Auto-MOEA/D-FP" and "Auto-MOEA/D-SS" are used to denote the obtained MOEA/D configurations in Table II (tuned under the final population framework) and Table III (tuned under the solution selection framework), respectively. Since Auto-MOEA/D-FP is tuned under the final population framework, it is expected that Auto-MOEA/D-FP has the best performance among all the algorithms. The Wilcoxon's rank sum test at a significant level of 5% is used to evaluate the statistical difference between Auto-MOEA/D-FP and each of the other six MOEA/D versions. The signs "+", "−", and "=" are used to indicate the compared MOEA/D version is statistically better than, worse than, or equivalent to Auto-MOEA/D-FP. The best and worst results among the seven MOEA/D versions are highlighted using yellow color and red color, respectively.

TABLE IV.
THE MEAN HYPERVOLUME VALUE OVER 31 RUNS OF EACH MOEA/D ON THREE-OBJECTIVE TEST PROBLEMS USING 10,000 SOLUTION EVALUATIONS UNDER THE FINAL POPULATION FRAMEWORK. THE BEST RESULT IS HIGHLIGHTED BY YELLOW COLOUR AND THE WORST RESULT IS HIGHLIGHTED BY RED FONT.

| Test problems ($M$=3) | Mean Hypervolume (10,000 solution evaluations) | | | | | | |
|---|---|---|---|---|---|---|---|
| | MOEA/D-WS | MOEA/D-TCH | MOEA/D-PBI | MOEA/D-IPBI | MOEA/D-MTCH | Auto-MOEA/D-FP | Auto-MOEA/D-SS |
| DTLZ1 | 0.1969 − | 0.6183 − | 0.4582 − | 0.0389 − | 0.6418 − | 0.6615 | 0.6712 = |
| DTLZ2 | 0.2487 − | 0.5303 − | 0.5552 − | 0.2487 − | 0.5578 = | 0.5585 | 0.2485 − |
| DTLZ3 | 0.0056 = | 0.0000 − | 0.0000 − | 0.0000 − | 0.0000 − | 0.0000 | 0.0000 − |
| DTLZ4 | 0.1275 − | 0.2750 − | 0.3295 − | 0.1775 − | 0.3618 − | 0.5573 | 0.5519 − |
| WFG1 | 0.4069 − | 0.4503 − | 0.3473 − | 0.4674 − | 0.4768 − | 0.5300 | 0.5444 = |
| WFG2 | 0.7586 − | 0.7819 − | 0.7537 − | 0.8012 − | 0.7922 − | 0.8969 | 0.8656 − |
| WFG3[1] | 0.2389 − | 0.4852 − | 0.4346 − | 0.4141 − | 0.4861 − | 0.5146 | 0.2265 − |
| WFG4 | 0.2488 − | 0.4921 − | 0.4573 − | 0.4535 − | 0.4948 − | 0.5148 | 0.5258 + |
| WFG5 | 0.2072 − | 0.4637 − | 0.4421 − | 0.4200 − | 0.4616 − | 0.4974 | 0.4927 − |
| WFG6 | 0.2135 − | 0.4692 − | 0.4376 − | 0.3426 − | 0.4702 − | 0.4939 | 0.2307 − |
| WFG7 | 0.2480 − | 0.5027 − | 0.3655 − | 0.4194 − | 0.4975 − | 0.5403 | 0.5291 − |
| WFG8 | 0.1738 − | 0.4416 − | 0.4063 − | 0.1863 − | 0.4361 − | 0.4675 | 0.4087 − |
| WFG9 | 0.1711 − | 0.4484 − | 0.3582 − | 0.2999 − | 0.4458 − | 0.4663 | 0.0825 − |
| Minus-DTLZ1 | 0.0279 − | 0.1953 − | 0.1861 − | 0.1309 − | 0.1915 − | 0.2196 | 0.2159 − |
| Minus-DTLZ2 | 0.5304 − | 0.5161 − | 0.5193 − | 0.5304 − | 0.5030 − | 0.5371 | 0.3714 − |
| Minus-DTLZ3 | 0.5010 − | 0.4772 − | 0.4715 − | 0.4951 − | 0.4690 − | 0.5252 | 0.5343 + |
| Minus-DTLZ4 | 0.4710 − | 0.5152 − | 0.3849 − | 0.4560 − | 0.5030 − | 0.5374 | 0.5290 − |
| Minus-WFG1 | 0.0190 − | 0.0664 − | 0.0498 − | 0.0318 − | 0.0902 − | 0.1083 | 0.0991 − |
| Minus-WFG2 | 0.1548 − | 0.2816 − | 0.2614 − | 0.2058 − | 0.2786 − | 0.2824 | 0.2634 − |
| Minus-WFG3 | 0.0240 − | 0.1848 − | 0.1886 − | 0.0742 − | 0.1818 − | 0.2160 | 0.1880 − |
| Minus-WFG4 | 0.5113 − | 0.4728 − | 0.4906 − | 0.5120 − | 0.4648 − | 0.5281 | 0.5251 − |
| Minus-WFG5 | 0.5108 − | 0.4926 − | 0.5029 − | 0.5113 − | 0.4745 − | 0.5346 | 0.5352 + |
| Minus-WFG6 | 0.5141 − | 0.4999 − | 0.5078 − | 0.5167 − | 0.4770 − | 0.5292 | 0.5379 + |
| Minus-WFG7 | 0.5136 − | 0.4624 − | 0.4835 − | 0.5146 − | 0.4557 − | 0.5288 | 0.0950 − |
| Minus-WFG8 | 0.5154 − | 0.5028 − | 0.5126 − | 0.5179 − | 0.4789 − | 0.5403 | 0.5233 − |
| Minus-WFG9 | 0.5057 − | 0.4780 − | 0.4875 − | 0.5071 − | 0.4559 − | 0.5217 | 0.5220 = |
| +/−/= | 0/25/1 | 0/25/1 | 0/25/1 | 0/25/1 | 0/23/3 | | 4/18/4 |

[1] Since the reference point set for WFG3 on the PlatEMO does not cover the flag region (the true Pareto front of the three-objective WFG3 has a flag region [38]), we generated the reference point set for the three-objective WFG3 problem by choosing non-dominated solutions from solution sets obtained by different EMO algorithms. We used NSGA-II, NSGA-III, MOEA/D-PBI, SMS-EMOA and SPEA2 with the population size 100 and 100,000 solution evaluations over 31 runs. A total of 7905 non-dominated solutions were obtained, which were used as the reference points for HV and IGD calculation of WFG3.

In Table IV, Auto-MOEA/D-FP shows high performance. Auto-MOEA/D-FP outperforms all the other six versions on almost all test problems. For some test problems, Auto-MOEA/D-FP is not the best in Table IV. However, the results obtained by Auto-MOEA/D-FP are very similar to the best results on those problems. From the experimental results, we can also see that the best results for almost all test problems are obtained by Auto-MOEA/D-FP and Auto-MOEA/D-SS. This observation shows that a tuning procedure is beneficial for improving the performance of MOEA/D.

Even though Auto-MOEA/D-SS shows the best performance on some test problems, it also shows the worst performance on some other test problems. For example, it has the worst performance on WFG3. However, Auto-MOEA/D-SS has the best performance when it is evaluated under the solution selection framework (which will be shown in Table VI in the next subsection). This observation suggests that the best algorithm configuration for the solution selection framework can be totally different from that for the final population framework.

The solution sets obtained by Auto-MOEA/D-FP and Auto-MOEA/D-SS for WFG3 under the evaluation of the final population framework are shown in Fig. 1. That is, Fig. 1 shows the solutions in the final population of a single run of each algorithm. A single run (over 31 runs) with the median HV value (from Table IV) is selected. We can see that the final population of Auto-MOEA/D-SS is not a good solution set. That is, only a few solutions are obtained since many solutions are overlapping with each other.

Fig. 2 shows the solution sets obtained by Auto-MOEA/D-FP and Auto-MOEA/D-SS for WFG3 under the evaluation of the solution selection framework. That is, Fig. 2 shows the selected solutions from all the examined solutions in the same single run as in Fig.1. We can see from the comparison between Fig. 1 and Fig. 2 that better solution sets can be obtained from the solution selection framework. We can also see that the obtained solution set by the solution selection framework in Fig. 2 (b) is the best with respect to the HV indicator among the four solution sets in Fig. 1 and Fig. 2, whereas the final population of the corresponding run in Fig. 1(b) is the worst.

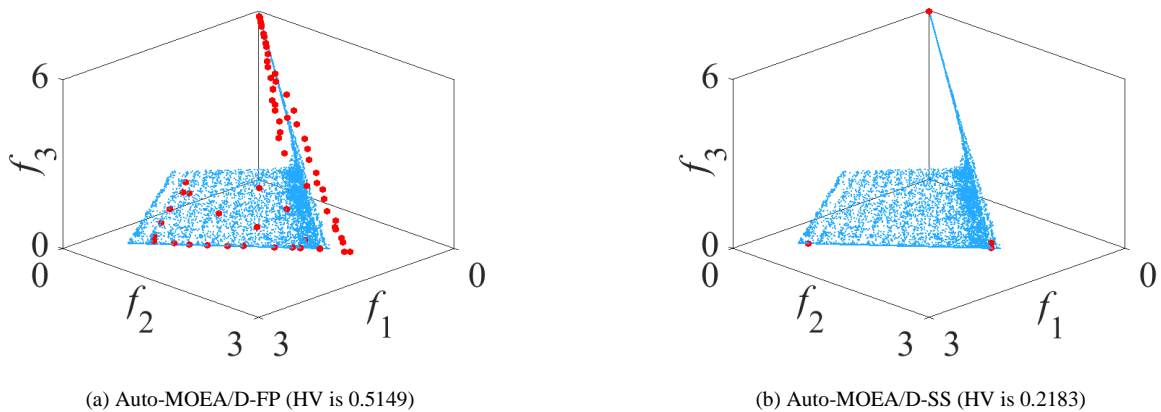

(a) Auto-MOEA/D-FP (HV is 0.5149)  (b) Auto-MOEA/D-SS (HV is 0.2183)

Figure 1. Solutions in the final population of a single run with the median HV value for WFG3 under the evaluation of the final population framework (with the stopping condition of 10,000 solution evaluations). The blue points are the Pareto front and the red points are the obtained solutions.

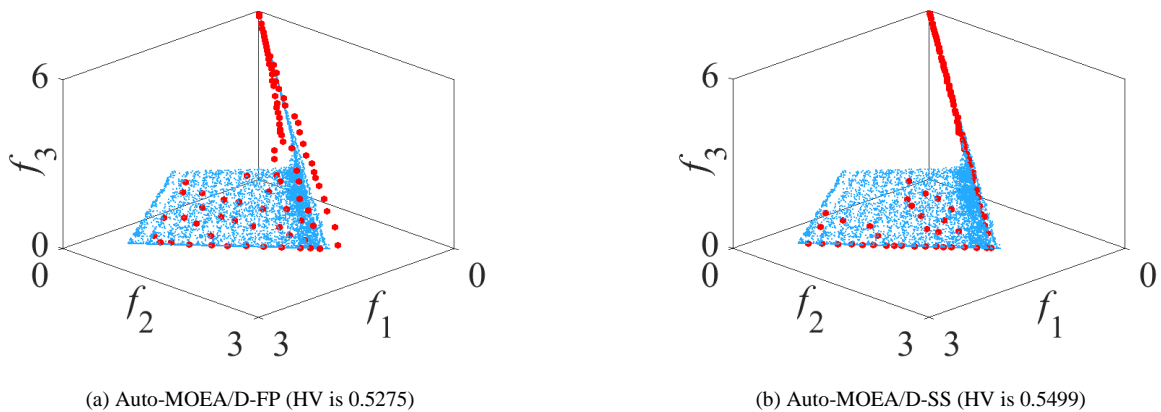

(a) Auto-MOEA/D-FP (HV is 0.5275)  (b) Auto-MOEA/D-SS (HV is 0.5499)

Figure 2. Selected solutions from all the examined solutions in a single run (corresponding to the same single run in Fig. 1) for WFG3 under the evaluation of the solution selection framework (with the stopping condition of 10,000 solution evaluations). The blue points are the Pareto front and the red points are the obtained solutions.

The performance of MOEA/D is also examined using a different performance indicator (i.e., IGD) and a different termination condition (i.e., 50,000 solution evaluations) under the final population framework. Table V shows the summary of statistical comparison results between Auto-MOEA/D-FP with the other six MOEA/D versions using the Wilcoxon rank sum test. The signs "+", "−", and "=" indicate the number of test instances on which the results of Auto-MOEA/D-FP are significantly better than, worse than, or equivalent to other MOEA/D versions.

We can see from Table V that Auto-MOEA/D-FP (which is designed for HV maximization under 10,000 solution evaluations) also shows high performance under different conditions. As an example, when Auto-MOEA/D-FP is evaluated using the HV indicator with 50,000 solution evaluations, its performance is significantly better than each of the other MOEA/D versions for at least 20 (out of 26) test problems. Auto-MOEA/D-FP also shows good performance when the IGD indicator is used for evaluation.

TABLE V.
SUMMARY OF STATISTICAL COMPARISON RESULTS BETWEEN AUTO-MOEA/D-FP WITH THE OTHER MOEA/D VERSIONS UNDER DIFFERENT CONDITIONS.

| Auto-MOEA/D-FP vs. | | MOEA/D-WS | MOEA/D-TCH | MOEA/D-PBI | MOEA/D-IPBI | MOEA/D-MTCH | Auto-MOEA/D-SS |
|---|---|---|---|---|---|---|---|
| Evaluation using HV with 50,000 solutions evaluations | + | 23 | 25 | 21 | 23 | 22 | 20 |
| | − | 3 | 1 | 4 | 3 | 3 | 5 |
| | = | 0 | 0 | 1 | 0 | 1 | 1 |
| Evaluation using IGD with 10,000 solutions evaluations | + | 25 | 23 | 23 | 25 | 22 | 19 |
| | − | 1 | 3 | 3 | 1 | 2 | 4 |
| | = | 0 | 0 | 0 | 0 | 2 | 3 |
| Evaluation using IGD with 50,000 solutions evaluations | + | 22 | 22 | 16 | 24 | 19 | 20 |
| | − | 3 | 4 | 7 | 2 | 6 | 4 |
| | = | 1 | 0 | 3 | 0 | 1 | 2 |

## C. EVALUATION UNDER THE SOLUTION SELECTION FRAMEWORK

In this subsection, each MOEA/D configuration is independently run 31 times on each test problem under the solution selection framework. The same parameter settings as in Section IV.B are used in the experiments. Actually, the same 31 runs are used in Section IV.B and IV.C. That is, all the examined solutions are stored in an unbounded external archive in each run in Section IV.B where only the final population was used. In this subsection, the output of each MOEA/D algorithm is a set of selected solutions from the unbounded external archive.

Various solution subset selection methods can be used to select pre-specified numbers of solutions from the unbounded external archive [19]. We use greedy HV-based and IGD-based solution subset selection methods in our experiments to evaluate each MOEA/D version in this paper. A recently proposed lazy greedy inclusion technique [39] is used for the speed-up of greedy inclusion solution subset selection. The lazy greedy HV-based inclusion method [39] is used in the solution selection framework when the performance of MOEA/D is evaluated by the HV indicator. Likewise, the lazy greedy IGD-based inclusion method is used when the performance of MOEA/D is evaluated by the IGD indicator. It should be noted that the distance-based greedy solution subset selection method is used for the design of Auto-MOEA/D-SS by the hyper-heuristic method. This is because the solution subset selection method needs to be performed 50,000 times in a single run of the hyper-heuristic method (i.e., a very fast method needs to be used).

Table VI shows the mean HV value over 31 runs of each MOEA/D version for each test problem under the solution selection framework. The termination condition in Table VI is 10,000 solution evaluations. The Wilcoxon's rank sum test at a significant level of 5% is used to evaluate the statistical difference between Auto-MOEA/D-SS and each of the six versions of MOEA/D. In Table VI, the signs "+", "−", and "=" are used to indicate that the compared MOEA/D version is statistically better than, worse than, or equivalent to Auto-MOEA/D-SS. The best and worst results among the seven MOEA/D versions are highlighted using yellow color and red color, respectively.

Auto-MOEA/D-SS shows high performance on many test problems when it is evaluated under the solution selection framework in Table VI. Even though Auto-MOEA/D-SS does not show the best performance on some test problems, it shows very similar performance to the best results obtained on those problems.

Whereas Auto-MOEA/D-SS shows the worst performance on DTLZ2, WFG3, WFG9, Minus-DTLZ2 and Minus-WFG7 in Table IV (evaluation under the final population framework), it shows the best performance on WFG3, WFG9, and Minus-WFG7, and comparable performance on DTLZ2 and Minus-DTLZ2 (i.e., the HV values are very similar to the best results on these two test problems) in Table VI. These observations clearly show that the optimal algorithm configuration for the solution selection framework can be totally different from that for the final population framework.

Moreover, we can see that better results are obtained in Table VI than Table IV for almost all cases. Actually, higher average HV values are obtained in 181 cases (out of 26 test problems × 7 algorithms = 182 cases). This observation shows the usefulness of the solution selection framework with an unbounded external archive. Theoretically, the result of the solution selection framework is always better than or equal to that of the final population framework if we can select the best solution set from the examined solutions. Since the HV-based greedy selection method is used (i.e., since the greedy algorithm is not an exact optimization algorithm), better results are obtained in Table IV than Table VI for one case (among the 182 cases). This observation suggests that further improvement is needed for solution selection in future studies.

TABLE VI.
THE MEAN HYPERVOLUME VALUE OVER 31 RUNS OF EACH MOEA/D ON THREE-OBJECTIVE TEST PROBLEMS USING 10,000 SOLUTION EVALUATIONS UNDER THE SOLUTION SELECTION FRAMEWORK (WITH THE LAZY GREEDY HV-BASED SOLUTION SUBSET SELECTION METHOD). THE BEST RESULT IS HIGHLIGHTED BY YELLOW COLOUR AND THE WORST RESULT IS HIGHLIGHTED BY RED FONT.

| Test problems (M=3) | 10,000 solution evaluations (HV) | | | | | | |
|---|---|---|---|---|---|---|---|
| | MOEA/D-WS | MOEA/D-TCH | MOEA/D-PBI | MOEA/D-IPBI | MOEA/D-MTCH | Auto-MOEA/D-FP | Auto-MOEA/D-SS |
| DTLZ1 | 0.4453 − | 0.6455 = | 0.4629 − | 0.1016 − | 0.6528 − | 0.6688 = | **0.6950** |
| DTLZ2 | 0.3917 − | 0.5658 + | 0.5629 − | 0.3919 − | 0.5655 + | **0.5658 +** | 0.5634 |
| DTLZ3 | **0.0065 =** | 0.0000 − | 0.0000 − | 0.0000 − | 0.0000 − | 0.0000 − | 0.0000 |
| DTLZ4 | 0.2023 − | 0.2821 − | 0.3345 − | 0.3503 − | 0.3670 − | 0.5652 − | **0.5660** |
| WFG1 | 0.4444 − | 0.4911 − | 0.3743 − | 0.4944 − | 0.5162 − | 0.5490 = | **0.5638** |
| WFG2 | 0.7926 − | 0.8275 − | 0.7683 − | 0.8259 − | 0.8339 − | **0.9068 +** | 0.8978 |
| WFG3 | 0.4367 − | 0.5103 − | 0.4565 − | 0.4863 − | 0.5169 − | 0.5267 − | **0.5431** |
| WFG4 | 0.5241 − | 0.5427 − | 0.4813 − | 0.5407 − | 0.5451 − | 0.5425 − | **0.5617** |
| WFG5 | 0.4866 − | 0.5185 − | 0.4628 − | 0.5041 − | 0.5180 − | 0.5172 − | **0.5210** |
| WFG6 | 0.3383 − | 0.5178 + | 0.4596 − | 0.3959 − | 0.5158 = | **0.5242 =** | 0.5144 |
| WFG7 | 0.4224 − | 0.5549 − | 0.4010 − | 0.4799 − | 0.5485 − | **0.5560 +** | 0.5550 |
| WFG8 | 0.2842 − | 0.4800 + | 0.4269 − | 0.2944− | 0.4781 + | **0.4861 +** | 0.4626 |
| WFG9 | 0.4471 − | 0.4987 − | 0.3970 − | 0.4390 − | 0.5013 − | 0.4987 − | **0.5184** |
| Minus-DTLZ1 | 0.1563 − | 0.2131 − | 0.2156 − | 0.2051 − | 0.2176 − | 0.2194 − | **0.2210** |
| Minus-DTLZ2 | **0.5430 +** | 0.5400 + | 0.5401 + | 0.5428 + | 0.5409 − | 0.5416 + | 0.5399 |
| Minus-DTLZ3 | 0.5168 − | 0.4963 − | 0.4931 − | 0.5072 − | 0.5007 − | 0.5387 − | **0.5401** |
| Minus-DTLZ4 | 0.4838 − | 0.5407 − | 0.4061 − | 0.4682 − | **0.5418 +** | 0.5404 − | 0.5409 |
| Minus-WFG1 | 0.0386 − | 0.0732 − | 0.0553 − | 0.0520 − | 0.1102 − | 0.1127 − | **0.1145** |
| Minus-WFG2 | 0.2199 − | 0.2948 − | 0.2787 − | 0.2386 − | 0.2956 − | 0.2925 − | **0.2965** |
| Minus-WFG3 | 0.1314 − | 0.2137 − | 0.2184 − | 0.1843 − | 0.2170 − | 0.2168 − | **0.2205** |
| Minus-WFG4 | 0.5389 = | 0.5078 − | 0.5208 − | 0.5390 = | 0.5237 − | **0.5417 +** | 0.5396 |
| Minus-WFG5 | 0.5365 − | 0.5235 − | 0.5292 − | 0.5362 − | 0.5297 − | 0.5382 − | **0.5393** |
| Minus-WFG6 | 0.5401 − | 0.5337 − | 0.5340 − | 0.5410 − | 0.5356 − | **0.5417 +** | 0.5415 |
| Minus-WFG7 | 0.5402 − | 0.4989 − | 0.5181 − | 0.5404 − | 0.5188 − | 0.5415 − | **0.5425** |
| Minus-WFG8 | 0.5419 − | 0.5388 − | 0.5388 − | 0.5425 + | 0.5392 − | **0.5427 +** | 0.5410 |
| Minus-WFG9 | 0.5328 − | 0.5150 − | 0.5208 − | 0.5334 − | 0.5178 − | 0.5354 = | **0.5355** |
| +/−/= | 1/22/3 | 4/20/2 | 1/24/1 | 2/22/2 | 3/19/4 | 8/12/6 | |

In Table VI, Auto-MOEA/D-SS shows high performance. However, if compared with the high performance of Auto-MOEA/D-FP in Table IV, the performance of Auto-MOEA/D-SS is not so dominant in Table VI. This is because the distance-based solution selection is used in the hyper-heuristic algorithm (for efficient calculation). If the HV-based solution selection method is used, the performance of Auto-MOEA/D-SS will be further improved. However, at the same time, much more computation time is needed.

In Table VI, we can also see that the difference in the average HV values between Auto-MOEA/D-FP and Auto-MOEA/D-SS is very small for all test problems (in comparison with the difference in Table IV) even when there exists a statistically significant difference. This may mean that the search for the best algorithm configuration is not easy under the solution selection framework since different configurations have similar performance. The search ability of the hyper-heuristic method will be improved by increasing the number of runs to evaluate each algorithm configuration. However, this needs more computation time.

The performance of MOEA/D under the solution selection framework is also examined using a different performance indicator (i.e., IGD) and a different termination condition (i.e., 50,000 solution evaluations). Table VII shows the summary of statistical comparison results between Auto-MOEA/D-SS with each of the other six MOEA/D versions using different evaluation conditions. The signs "+", "−", and "=" indicate the number of test instances on which the results of Auto-MOEA/D-SS are significantly better than, worse than, or equivalent to other MOEA/D versions.

In Table VII, Auto-MOEA/D-SS shows similar performance to Auto-MOEA/D-FP on average when the termination condition of 50,000 solution evaluations is used. Although statistical differences are observed in the performance between Auto-MOEA/D-SS and Auto-MOEA/D-FP, their average HV values are very similar as in Table VI. Auto-MOEA/D-SS outperformed the other five MOEA/D versions on average in Table VII when the HV indicator is used for performance comparison.

TABLE VII.
SUMMARY OF STATISTICAL COMPARISON RESULTS BETWEEN AUTO-MOEA/D-SS WITH THE OTHER MOEA/D VERSIONS UNDER DIFFERENT CONDITIONS.

| Auto-MOEA/D-SS vs. | | MOEA/D-WS | MOEA/D-TCH | MOEA/D-PBI | MOEA/D-IPBI | MOEA/D-MTCH | Auto-MOEA/D-FP |
|---|---|---|---|---|---|---|---|
| Evaluation using HV with 50,000 solutions evaluations | + | 19 | 17 | 19 | 20 | 16 | 9 |
| | − | 4 | 5 | 5 | 6 | 5 | 10 |
| | = | 3 | 4 | 2 | 0 | 5 | 7 |
| Evaluation using IGD with 10,000 solutions evaluations | + | 18 | 12 | 18 | 19 | 8 | 5 |
| | − | 6 | 8 | 6 | 7 | 8 | 16 |
| | = | 2 | 6 | 2 | 0 | 10 | 5 |
| Evaluation using IGD with 50,000 solutions evaluations | + | 19 | 12 | 9 | 19 | 7 | 6 |
| | − | 7 | 14 | 9 | 7 | 15 | 17 |
| | = | 0 | 0 | 8 | 0 | 4 | 3 |

When the IGD indicator is used in Table VII, the performance of Auto-MOEA/D-SS is not the best. For more than 15 test problems (out of 26 test problems), Auto-MOEA/D-SS is outperformed by Auto-MOEA/D-FP. One possible reason is that different configurations are needed to obtain good solution sets for different performance indicators. That is, the obtained configurations for the HV indicator are not always good for the IGD indicator. Another possible reason is that the optimization of Auto-MOEA/D-SS has not been fully completed as discussed for Table VI (i.e., use of the distance-based selection method and similar performance of different configurations).

## VI. CONCLUSIONS

In this paper, we empirically demonstrated the usefulness and flexibility of the solution selection framework using the MOEA/D algorithm on 26 test problems. An offline genetic algorithm-based hyper-heuristic method was used to search for the optimal MOEA/D configurations for each test problem under the final population framework and the solution selection framework. The experimental results suggested the optimal configurations can be totally different under the two frameworks. Better solution sets were obtained from the solution selection framework with a greedy solution subset selection method for almost all test problems than the final population framework.

As shown in this paper (and some other studies [14], [19]), better solution sets are usually obtained from the solution selection framework than the final population framework. However, new algorithm design has not been studied under the solution selection framework in the literature. This is clearly a promising future research direction. Another important research direction is the development of a new efficient and effective solution subset selection method to facilitate the solution selection framework.

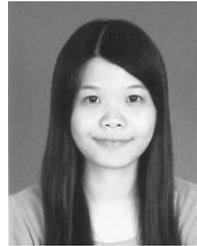

**LIE MENG PANG** (M'18) received her Bachelor of Engineering degree in Electronic and Telecommunication Engineering and Ph.D. degree in Electronic Engineering from the Faculty of Engineering, Universiti Malaysia Sarawak, in 2012 and 2018, respectively.

She is currently a postdoctoral researcher with the Department of Computer Science and Engineering, Southern University of Science and Technology (SUSTech), China. Her research interests include fuzzy systems and evolutionary computations.

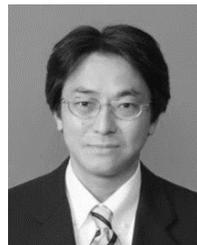

**HISAO ISHIBUCHI** (M'93–SM'10–F'14) received the B.S. and M.S. degrees in precision mechanics from Kyoto University, Kyoto, Japan, in 1985 and 1987, respectively, and the Ph.D. degree in computer science from Osaka Prefecture University, Sakai, Osaka, Japan, in 1992. He was with Osaka Prefecture University in 1987-2017. Since 2017, he is a Chair Professor at Southern University of Science and Technology, China. His research interests include fuzzy rule-based classifiers, evolutionary multi-objective and many-objective optimization, memetic algorithms, and evolutionary games.

Dr. Ishibuchi was the IEEE CIS Vice-President for Technical Activities in 2010-2013, an AdCom member of the IEEE CIS in 2014-2019, and the Editor-in-Chief of the IEEE COMPUTATIONAL INTELLIGENCE MAGAZINE in 2014-2019.

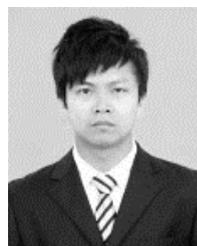

**KE SHANG** (M'19) received the B.S. degree and the Ph.D. degree from Xi'an Jiaotong University, China, in 2009 and 2016, respectively. From 2017 to 2019, he was a postdoc researcher at the Department of Computer Science and Engineering, Southern University of Science and Technology (SUSTech), China. He is currently a research assistant professor at the same department in SUSTech.

His current research interests include evolutionary multi-objective optimization and its applications. He is the recipient of the GECCO 2018 Best Paper Award and CEC 2019 First Runner-up Conference Paper Award.